\documentclass[letterpaper, 10 pt, conference]{ieeeconf}  % Comment this line out if you need a4paper

\IEEEoverridecommandlockouts                              % This command is only needed if 
\usepackage{hyperref}
\hypersetup{
    colorlinks=true,
    linkcolor=magenta,
    filecolor=magenta,      
    urlcolor=magenta,
    pdftitle={Surgical Gym: A high-performance GPU-based platform for reinforcement leanring with surgical robots},
    pdfpagemode=FullScreen,
    }
\usepackage{graphicx}
\usepackage{float}% http://ctan.org/pkg/multicols

\usepackage{amsmath,amssymb,amsfonts}
    
\title{\LARGE \bf
Surgical Gym: A high-performance GPU-based platform for reinforcement learning with surgical robots
}

\author{Samuel Schmidgall$^{1}$, Axel Krieger$^{2}$ and Jason Eshraghian$^{3}$% <-this % stops a space
\thanks{$^{1}$Samuel Schmidgall is with the Department of Electrical Engineering, Johns Hopkins University, Baltimore, MD, USA
        {\tt\small sschmi46@jhu.edu}}%
\thanks{$^{2}$Axel Kriger is with the Department of Mechanical Engineering, Johns Hopkins University,
        Baltimore, MD, USA
        {\tt\small axel@jhu.edu}}%
\thanks{$^{2}$Jason Eshraghian is with the Department of Electrical Engineering, University of California, Santa Cruz,
        Santa Cruz, CA, USA
        {\tt\small jeshragh@ucsc.edu}}%
}

\begin{document}

\maketitle
\thispagestyle{empty}
\pagestyle{empty}

%%%%%%%%%%%%%%%%%%%%%%%%%%%%%%%%%%%%%%%%%%%%%%%%%%%%%%%%%%%%%%%%%%%%%%%%%%%%%%%%
\begin{abstract}

Recent advances in robot-assisted surgery have resulted in progressively more precise, efficient, and minimally invasive procedures, sparking a new era of robotic surgical intervention. 
This enables doctors, in collaborative interaction with robots, to perform traditional or minimally invasive surgeries with improved outcomes through smaller incisions.
Recent efforts are working toward making robotic surgery more autonomous which has the potential to reduce variability of surgical outcomes and reduce complication rates.
Deep reinforcement learning methodologies offer scalable solutions for surgical automation, but their effectiveness relies on extensive data acquisition due to the absence of prior knowledge in successfully accomplishing tasks. %#3
Due to the intensive nature of simulated data collection, previous works have focused on making existing algorithms more efficient. %#4
In this work, we focus on making the simulator more efficient, making training data much more accessible than previously possible. %#5
We introduce Surgical Gym, an \textit{open-source} high performance platform for surgical robot learning where both the physics simulation and reinforcement learning occur directly on the GPU. %#6
We demonstrate between \textit{100-5000}$\times$ faster training times compared with previous surgical learning platforms. %#7
The code is available at: \hyperlink{https://github.com/SamuelSchmidgall/SurgicalGym}{https://github.com/SamuelSchmidgall/SurgicalGym.}

\end{abstract}

% Intro, Prior Art, Methods, Experimental Design, Results, Discussion, Conclusion.

%%%%%%%%%%%%%%%%%%%%%%%%%%%%%%%%%%%%%%%%%%%%%%%%%%%%%%%%%%%%%%%%%%%%%%%%%%%%%%%%
\section{INTRODUCTION}

The field of surgery has undergone a major transformation with the introduction of robots. These robots, controlled directly by surgeons, have enhanced the precision and efficiency of complex surgical procedures \cite{attanasio2021autonomy}. In practice, the surgeon operates the robot from a control station using tools like hand controllers and foot pedals. The robot, in response, carries out the surgeon's movements with great accuracy, eliminating even the smallest amount of shaking.

In modern medical practice, surgical robots play a pivotal role in supporting surgeons during minimally invasive procedures, with over a million surgeries per year carried out by human surgeons guiding these robots \cite{tindera2019robot}. Despite their complexity, surgical robots in practice are not autonomous and rely on human control to function. The incorporation of autonomy has the potential to dramatically improve surgical outcomes since automation of these tasks can alleviate the burden of repetitive tasks and minimize surgeon fatigue \cite{haidegger2019autonomy}. The growing desire to improve surgical efficiency has led to a growth in efforts focused on automating various surgical tasks, ranging from suturing \cite{barnoy2021robotic, chiu2021bimanual, varier2020collaborative}, endoscope control \cite{pore2022colonoscopy, turan2019learning, trovato2010development}, and tissue manipulation \cite{li2020super, shin2019autonomous, pedram2020toward, lu2021super}, to pattern cutting \cite{nguyen2019new, thananjeyan2017multilateral, nguyen2019manipulating}.

\begin{figure}
    \centering
    \includegraphics[width=0.99\linewidth]{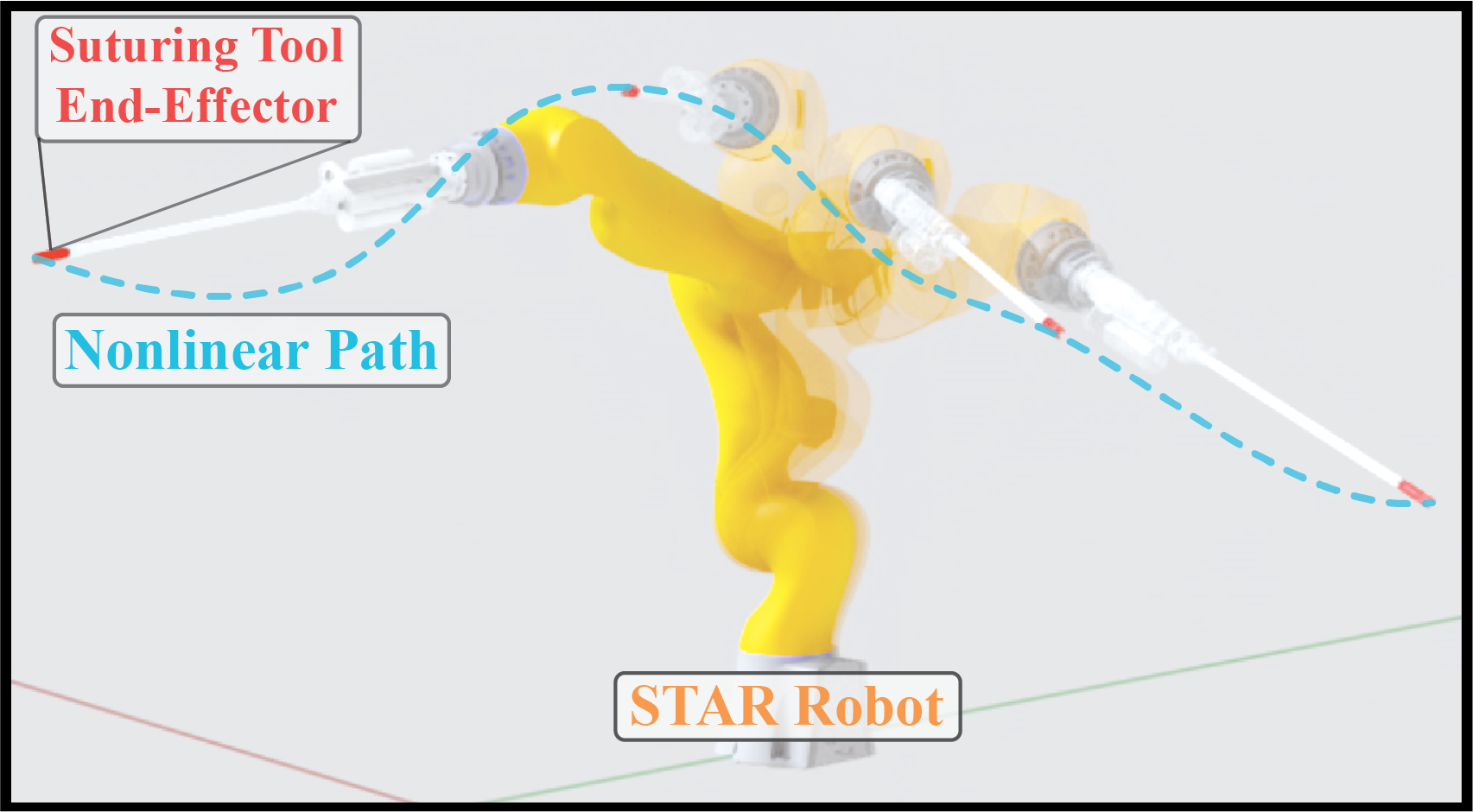}
    \caption{Demonstration of Smart Tissue Autonomous Robot robot performing path following (path in blue dashed lines) going from right to left. Suturing tool end-effector (red) precisely follows randomly generated points along a nonlinear path.}
    \label{fig:surgical-envs}
\end{figure}

\begin{figure*}
    \centering
    \includegraphics[width=0.99\linewidth]{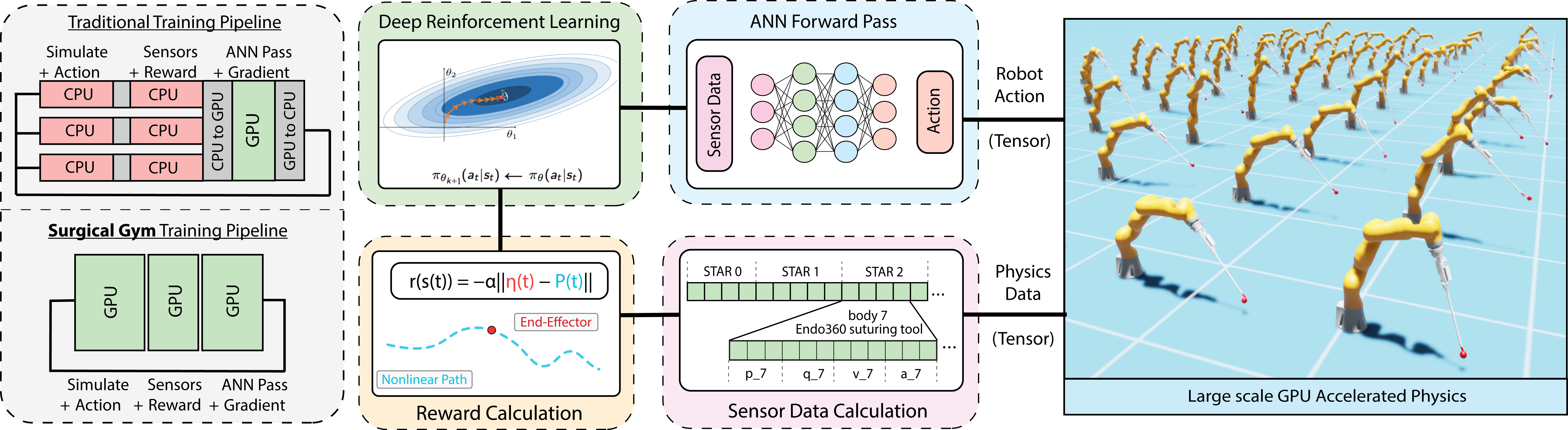}
    \caption{\textbf{(Left) Data process flow comparison.} Comparison between the traditional reinforcement learning pipeline which use CPU based physics engines and Surgical Gym, which uses a GPU based physics engine. Surgical Gym avoids expensive data transfer between the CPU and GPU, with all operations tensorized on the GPU. \textbf{(Right) Detailed Surgical Gym process flow.} Large scale GPU accelerated physics are executed with thousands of robots in parallel. Sensor data is calculated from actor bodies and their corresponding positions, rotations and velocities which are stored directly in PyTorch tensors. Reward calculation takes relevant sensor data, calculates a per-agent reward, which is passed to the GPU-accelerated Proximal Policy Optimization implementation for gradient calculation. Finally, sensor data is forward propagated through an ANN to calculate an action, which is sent to the physics engine.
}
    \label{fig:main-figure}
\end{figure*}

Recently, deep reinforcement learning (RL) methods have shown they can produce various control strategies exceedingly well, which is promising for automated surgery~\cite{yip2023artificial, nguyen2019manipulating, scheikl2022sim, barnoy2022control, ou2023sim}. However, this development does not come without a myriad of challenges. One of the most significant challenges results from the fact that RL-based methods often require a large amount of data to succeed, particularly because they start without any pre-existing domain knowledge \cite{pertsch2021guided, xu2021surrol, huang2023guided}. % ADD CITATIONS
While strategies using specialized domain knowledge provide significant value (\textit{e.g. expert demonstration from surgeon}), there remains a substantial need for larger and more diverse training datasets to improve task generalization \cite{heess2017emergence, hilton2023scaling}. 
Building on this, the field of robotics has recently experienced a surge of major advancements with learning-based control algorithms due to the development of high-performance GPU-based physics simulators, which provide \textit{100-10000}$\times$ faster data collection rates than CPU based simulators \cite{makoviychuk2021isaac, radosavovic2023real, chen2022system, qi2023hand}.

Despite these advancements in other areas of robotics, the potential benefits of these high-performance simulators for surgical robot learning are yet to be realized. We believe that providing easy access to surgical robotic training data is the next step toward building more capable and autonomous surgical robots. Toward this, we present \textbf{Surgical Gym}, an open-source GPU-based learning platform for surgical robotics.

Our main contributions are as follows:

\begin{itemize}
    \item We introduce an open-source high-performance platform for surgical robot learning that accelerates training times up to \textit{5000}$\times$ by leveraging GPU-based physics simulation and reinforcement learning, thereby making surgical robot training data more accessible.
    \item We built \textit{five} training environments which support learning for \textit{six} different surgical robots \& attachments (da Vinci Research Kit \cite{kazanzides2014open} and the Smart Tissue Autonomous Robot \cite{shademan2016supervised,saeidi2022autonomous}). GPU-based RL algorithms are incorporated into this library, making training on these environments simple and reproducible even without RL expertise. %with \textit{four} unique surgical robot attachments
\end{itemize}

 We believe that providing easy access to surgical robot training data is the next step toward building surgical robots with greater autonomy, and hope Surgical Gym provides a step forward in this direction. All of the code and environments are \textit{open-source} and can be found at the following link: \hyperlink{https://github.com/SamuelSchmidgall/SurgicalGym}{https://github.com/SamuelSchmidgall/SurgicalGym.}

\section{Background}

\subsection*{Reinforcement Learning for Robotics}

RL is a machine learning paradigm where an agent learns from the environment by interacting with it and receiving rewards for performing actions \cite{sutton2018reinforcement}. The fundamental concept behind reinforcement learning is trial-and-error search, i.e., the agent selects an action $a_t$ when given a particular state $s_t$, receives a reward $r_t$ based on that state, and then transitions to a new state $s_{t+1}$. The goal of the agent is to maximize the expected cumulative reward over time. Mathematically, the environment is modeled as a Markov Decision Process (MDP), defined by a tuple $(S, A, P, R)$, where $S$ is the space of possible states, $A$ is the space of possible actions, $P$ is the state transition probability, i.e., $P(s'|s,a)$ is the probability of transitioning to state $s'$ after taking action $a$ in state $s$, and $R$ is the reward function. $R(s,a)$ is the expected reward received after taking action $a$ in state $s$.

The agent's behavior is defined by a policy ($\pi$), which is a mapping from states to probabilities of selecting each action, i.e., $\pi(a|s)$ is the probability of taking action $a$ in state $s$. The value of a state, denoted by $V^\pi(s)$, is the expected cumulative reward when starting from state $s$ and following policy $\pi$. The Bellman equations for $V^\pi(s)$ is defined as:

\[ V^\pi(s) = \sum_{a} \pi(a|s) \left[R(s,a) + \gamma \sum_{s'} P(s'|s,a) V^\pi(s')\right] \]

\noindent where $\gamma$ is the discount factor, $s'$ is the next state, and $a'$ is the next action.

The objective in reinforcement learning is to find the optimal policy $\pi^*$ that maximizes the expected cumulative reward, i.e., $\pi^* = \arg\max_{\pi} V^\pi(s)$.

\subsection*{Surgical Robot Algorithms}

Previous efforts in surgical robotics have focused on developing sophisticated controllers for particular subtasks \cite{palep2009robotic,leonard2014smart, shademan2016supervised, saeidi2022autonomous}. While these approaches are quite successful and reliable, they typically focus on solving surgical sub-problems independently. Considering the wide diversity of scenarios that clinical surgeries may present, it is unlikely that isolated subtask solutions will lead to full surgical \textit{automation}. 

RL, on the other hand, demonstrates the ability to generalize well to novel circumstances given a sufficient amount of experience \cite{packer2018assessing}. However, while control algorithms have been demonstrated to be reliable over decades of research, RL algorithms are relatively new and have yet to be deployed in patient-critical applications. This is particularly important for high-precision problems like automated surgery. Therefore, it is likely a combination of control-based and RL-based approaches will enable highly reliable and autonomous surgery.

\subsection*{Surgical Robot Platforms}

\paragraph*{The da Vinci}  The da Vinci Surgical System is a robotic surgical system designed to facilitate complex surgery using a minimally invasive approach \cite{dimaio2011vinci}. 

The system operates through an interface with the Master Tool Manipulator (MTM), which serves as the control center for the surgeon to direct surgical actions. Through the MTM's handles or joysticks, the surgeon's movements are translated into corresponding motions of the \textbf{Patient Side Manipulators (PSMs)}, which are robotic arm attachments responsible for performing the surgery. The PSMs are flexible, multi-jointed instruments capable of holding and manipulating surgical tools, adjusting to the unique anatomy and requirements of each procedure. 

A third component of the da Vinci system is the \textbf{Endoscopic Camera Manipulator (ECM)}. The ECM is another robotic arm attachment that holds and controls the movement of a stereo endoscope, a special camera that provides a high-definition, three-dimensional view of the surgical field. This allows the surgeon, from the control console, to have a detailed and magnified view of the area being operated on, improving precision during the surgical procedure.

\paragraph*{The Smart Tissue Autonomous Robot} The \textbf{Smart Tissue Autonomous Robot (STAR)} \cite{leonard2014smart, shademan2016supervised} is an autonomous robot designed to perform soft tissue surgery with minimal assistance from a surgeon. The STAR robot has been used for a variety of autonomous surgical procedures, most notably, the first autonomous laparoscopic surgery for intestinal anastomosis \cite{saeidi2022autonomous} (reconnection of two tubular structures such as blood vessels or intestines). The STAR performed the procedure in four different animals, producing better results than humans executing the same procedure.

The STAR uses a seven DoF light-weight arm mounted with an actuated suturing tool, which has actuators to drive a circular needle and a pitch axis (see \textit{Robot Descriptions}). The STAR supports a manual mode and an automatic mode. In \textit{manual mode}, a surgeon selects where each stitch is placed, whereas in \textit{automatic mode}, an incision area is outlined and the STAR determines where each stitch is placed. Automatic mode chooses these placements based on the contour of the incision, which, once chosen, is transformed to the 3D camera frame and stitches are distributed evenly \cite{leonard2014smart}.

\subsection*{Software Platforms for Surgical Robot Learning}

The first open-source RL-focused environment for surgical robotics is the da Vinci Reinforcement Learning (dVRL) suite \cite{richter2019open}, which aims to enable scientists without a surgical background to develop algorithms for autonomous surgery. dVRL introduces two straightforward training tasks: a robotic reach and a pick-and-place problem. Transfer to robotic hardware is demonstrated using the robotic target reaching policy for suctioning blood using a simulated abdomen (created by molding pig parts into gelatin and then filled with blood). Transfer to hardware was made possible since environments were designed around being compatible with the da Vinci Research Kit (dVRK) \cite{kazanzides2014open}, a large-scale effort which enables institutions to share of a common hardware platform for the da Vinci system.

UnityFlexML \cite{tagliabue2020unityflexml} presents a framework that expands on autonomous surgical training, focusing on manipulating deformable tissues during a nephrectomy procedure. This framework was built on Unity3D, which naturally supports deformable objects. Like dVRL, the policy in UnityFlexML demonstrates successful transferability to the dVRK.

LapGym \cite{scheikl2023lapgym} provides an RL environment for developing and testing automation algorithms in laparoscopic surgery, covering four main skills: spatial reasoning, deformable object manipulation \& grasping, dissection, and thread manipulation. This includes a rich set of tasks from rope threading to tissue dissection. Unlike previous simulators, it presents a variety of image-based learning problems from which it supports RGB, depth, point clouds and semantic segmentation.

Finally, SurRoL \cite{xu2021surrol} is an open-source RL focused simulation platform, like previous work, designed to be compatible with the dVRK. SurRoL develops environments for \textit{both} da Vinci attachments (patient side and endoscopic camera manipulators, see subsection \textit{The da Vinci System}) as well as for multiple arms. 

SurRoL and LapGym have many environments, where SurRoL
implements \textit{ten} unique training tasks and LapGym implements a total of \textit{twelve}. Other platforms only supported a limited number of training tasks, with UnityFlexML having \textit{one} and dVRL having \textit{two}. %and AMBF-RL 
Applications of RL on these environments focus on data-efficiency, aiming to develop algorithms which use few environment interactions, e.g. with SurRoL only 100,000 \cite{xu2021surrol, huang2023guided} interactions in their experiments. This is because data collection in these environments is \textbf{\textit{expensive}} (see subsection \textit{Simulator Performance Comparison}) and thus experimentation is much more limited.

% SurRoL 

\begin{figure}
    \centering
    \includegraphics[width=0.95\linewidth]{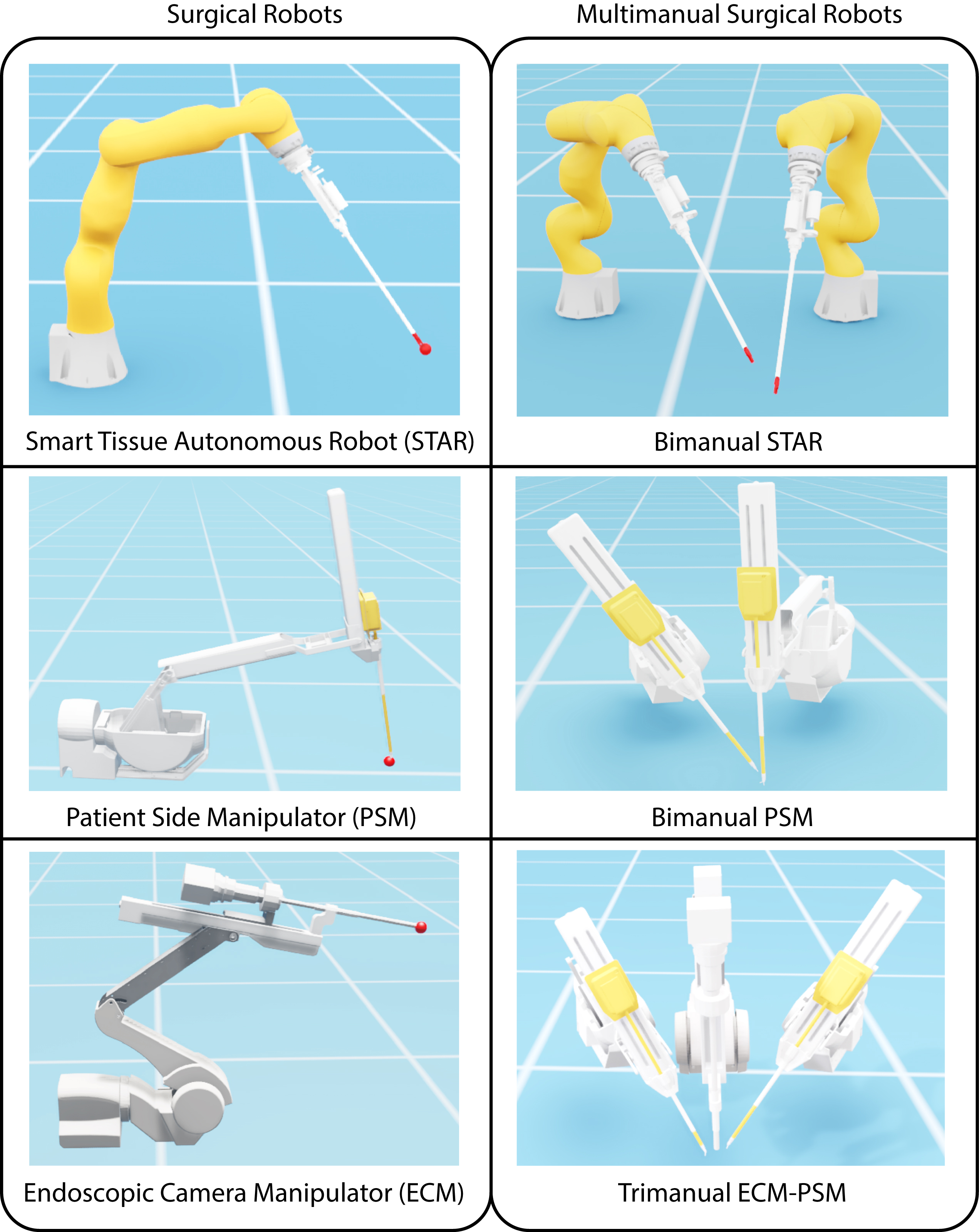}
    \caption{Demonstration of surgical robots \& attachments supported in Surgical Gym on a target reaching task. }
    \label{fig:surgical-envs}
\end{figure}

\section{Methods}

\subsection*{Simulation Setup}

We build Surgical Gym on Isaac Gym \cite{makoviychuk2021isaac}, which is a tensorized physics platform utilizing PhysX. The physics simulation as well as the observation, reward, and action calculation takes place on the GPU, enabling direct data transfer from the physics buffers to GPU tensors in PyTorch \cite{paszke2019pytorch}. This process bypasses any potential slowdowns that could occur if the data were to pass through the CPU. 

In Surgical Gym, an actor is composed of rigid bodies connected by joints. Rigid bodies are the primitive shapes or meshes that comprise of the actor. Different joint types link rigid bodies, each having a specific number of DoFs: fixed joints have 0, revolute and prismatic joints have 1, and spherical joints have 3. Joints can take different types of simulation input, including forces, torques, and PD controls (position or velocity targets). The action space for Surgical Gym environments are currently configured for position-based (PD) commands. However, each environment also supports torque and velocity-based commands. %

Like experiments in Isaac Gym, we use the Temporal Gauss Seidel (TGS) solver \cite{macklin2019small} to calculate future object states. It uses sub-stepping to accelerate convergence better than solvers using larger steps with multiple iterations. The solver computes and accumulates scaled velocities into a delta buffer per iteration, which is then projected onto constraint Jacobians and added to the biases in constraints. This process has a similar computational cost as the traditional Gauss-Seidel solver \cite{bagnara1995unified}, but without the expense of sub-stepping. For positional joint constraints, an extra rotational term is calculated for joint anchors to prevent linearization artifacts.

\subsection*{Robot Descriptions}

To support dVRK compatibility for the PSM and ECM attachments, URDF descriptor models were derived from the dVRK-ros\footnote{https://github.com/jhu-dvrk/dvrk-ros} speficiations \cite{kazanzides2014open} and converted to Universal Scene Descriptor (USD) format to be compatible with the GPU-based physics engine. For the STAR robot, we used descriptors for the Kuka LWR (7 DoF) arm and the suturing tool from the Endo360 specification, as was used in suturing demonstrations with the STAR robot \cite{saeidi2022autonomous}.

The PSM has seven DoFs, where the arm position is determined by the first six (denoted as $\theta_i$), and the seventh DoF corresponds to the PSM jaw angle, which, from an action perspective, is either open or closed ($a_{jaw} \in \{0, 1\}$) but has an intermediate angle ($\theta_{jaw} \in \mathbb{R}$) during the process of opening or closing. The PSM arm includes both revolute (R) and prismatic (P) actuated joints, arranged in the following sequence: RRPRRR.

The ECM tool follows a similar structure to the PSM, however it does not have a jaw, rather, a camera attached to the tool tip. The ECM arm also consists of revolute and prismatic joints arranged in the following sequence: RRPRRR.

The STAR robot has 8 DoFs with a suturing tool attachment extending from the end of the robot arm. The STAR arm consists entirely of revolute joints arranged in the following sequence: RRRRRRRR. 

\paragraph*{\textbf{Action Space}}

Although the dVRK supports 6-DoF motion (and the STAR 7-DoF), previous works have defined the action space as positional coordinates of the end-effector with a dynamics controller actuating the joints \cite{xu2021surrol, scheikl2023lapgym}. While this accelerates learning where data is scarce, (1) it can restrict the utilization of the robot's complete motion capabilities and (2) it tends to be computationally expensive. However, in practice, end-effector control is likely a much better option from a safety perspective. We opted to define the action space in a way that enables the full range of robot motions via torque or PD control, capitalizing on the extensive data available in Surgical Gym, with the opportunity for custom controllers to be wrapped on top of this action space for all of the supported robots.

\paragraph*{\textbf{Observation Space}}
While the observation space differs slightly between tasks, most of it remains consistent. The base observation space for the robot is defined by the concatenation of  DoF positions ($\boldsymbol{\theta} \in \mathbb{R}^{n}$ where $n = $ number of DoFs), DoF velocities ($\dot{\boldsymbol{\theta}} \in \mathbb{R}^{n}$), the tool link tip position ($\boldsymbol{p}_{tip} \in \mathbb{R}^{3}$), and the current PD-targets for the DoF ($\boldsymbol{\theta}_{target} \in \mathbb{R}^{n}$). With tasks involving a goal position (e.g. Target Reaching and Active Tracking) the goal position ($g$ or $g_t \in \mathbb{R}^{3}$) is included in the observation vector. Similarly, with tasks involving image input (e.g. Endoscopic Image Matching) the target image ($p_{x,y,target}$) and the current end-effector image ($p_{x,y}$) is provided. 

\subsection*{Policy learning and control}

For policy learning in Surgical Gym, we utilize a GPU-optimized implementation of Proximal Policy Optimization (PPO) algorithm \cite{schulman2017proximal} which has been adapted for efficient parallel learning with numerous robots \cite{rudin2022learning}. 

Previous work found that the hyper-parameter, \textit{batch size} $B = n_{\text{robots}}n_{\text{steps}}$, plays a significant role in learning efficiency. Here, $n_{\text{steps}}$ signifies the steps each robot takes per update, and $n_{\text{robots}}$ denotes parallel-simulated robots. By increasing $n_{\text{robots}}$, we decrease $n_{\text{steps}}$ to optimize training times. However, to ensure algorithm convergence and effective Generalized Advantage Estimation (GAE) \cite{schulman2015high}, $n_{\text{steps}}$ cannot be reduced below a threshold, which has been shown to be fewer than 25 steps or 0.5s of simulated time \cite{rudin2022learning}. For backpropagation, the batch is split into large mini-batches, which improves the learning stability without increasing training time.

During training, robot resets, triggered either by falls or after specific intervals to facilitate new trajectory exploration, disrupt the infinite reward horizon assumed by the PPO algorithm's critic function, potentially impairing performance. To rectify this, we differentiate between terminations due to falls or goal attainment, which the critic can anticipate, and time-outs, which are unpredictable. We address time-out cases by augmenting the reward with the critic's prediction of the infinite sum of future discounted rewards, a technique known as ``bootstrapping.'' \cite{rudin2022learning} This strategy, necessitating modification of the standard Gym interface to detect time-outs, effectively maintains the infinite horizon assumption.

\section{Results}

\subsection*{Environments}

Surgical Gym makes RL easier to use by including a common Gym-like interface \cite{brockman2016openai}, which helps make the process of developing and evaluating algorithms more straightforward. Included in this interface, we have developed a range of learning-based tasks that are useful for surgical automation, covering varying degrees of task complexity. We build five tasks that support five different surgical robots \& attachments. We chose to incorporate tasks that develop low-level control capabilities (e.g. end-effector target reaching, image-guided navigation) to build a foundation for more advanced problems that would require these skills. These five tasks are described in more detail below.

\begin{itemize}
    \item \textbf{Target Reaching:} The Target Reaching task involves aligning the end-effector, $\eta_{t}$, of the robot with a red target sphere with a randomly sampled position, $g$, for each environment. The task's reward structure is devised based on the distance between the end-effector and the sphere.
    \begin{equation*}
        r(s_t) = \rho ||\eta_{t} - g||
    \end{equation*}
    This environment supports the PSM and ECM attachments for the da Vinci, and the STAR robot with the suturing tool end-effector. The goal point for this task has a randomization range of $g_{x,y,z}\sim \mathcal{N}(0, 0.05)$ in front of the robot for the ECM \& PSM and $g_{x,y,z}\sim \mathbb{N}(0, 0.15)$ for the STAR, providing a wide reaching range relative to the robot.
    
    \item \textbf{Active Tracking:} In the Active Tracking task, similar to the Target Reaching task, the goal is to align the end-effector with a target position, $g_{t}$. Unlike the previous task, the goal location in this environment is time-varying and changes its position in all directions $(x, y, z)$ as follows:
    \begin{equation*}
        g_{t} = g_{t} + \dot{g}_{t}
    \end{equation*}
    \begin{equation*}
        \dot{g}_{t} = \dot{g}_{t} + \mathbb{N}(0, 1^{-4})
    \end{equation*}
    
    This environment supports the PSM and ECM attachments for the da Vinci, and the STAR robot. The task reward system is based on the distance between the moving sphere and the end-effector. The tracking problem samples the goal coordinates from $g_{x,y,z}\sim \mathcal{N}(0, 0.05)$ for ECM \& PSM and $g_{x,y,z}\sim \mathcal{N}(0, 0.15)$ for STAR with a max positional offset of $-0.2$ and $0.2$.
    
    %\begin{equation*}
    %    r(s_t) = \rho ||\eta_{t} - g_t||
    %\end{equation*}

    \item \textbf{Endoscopic Image Matching:} In the Endoscopic Image Matching task, the sensory input is a specific target image ($p_{x,y, target})$, and the goal is to attain the same image through the endoscopic camera of the ECM robot on its end-effector ($p_{x,y}$). The task is about creating an accurate match between the given image and the current viewpoint of the end-effector. The reward is defined as follows:
    
    \begin{equation*}
        r(s_t) = \frac{1}{w \cdot h} \sum_{x,y} (p_{x,y} - p_{x,y, target})
    \end{equation*}
    
    This task tests the robot's ability to precisely position and orient itself to match a specific visual scene, simulating the surgical requirement of aligning tools to match reference images.

    \item \textbf{Path following:} In the Path Following task, the robot's objective is to navigate its end-effector, $\eta_{t}$, along a predefined trajectory, $\mathcal{P}$, which is represented as a sequence of waypoints in the environment. The main challenge lies in maintaining a consistent alignment with the path while adapting to any minor perturbations in the environment. 
    The path waypoints are sampled using a parametric representation of the path, which is application configurable. By default, a cubic spline $\mathcal{S}(t)$ is used to represent the path $\mathcal{P}$, which interpolates the given waypoints. The cubic spline $\mathcal{S}(t)$ can be represented as:

    \[ \mathcal{S}(t) = a(t - t_{i})^3 + b(t - t_{i})^2 + c(t - t_{i}) + d, \]

    where $[a, b, c, d]$ are the coefficients of the spline, which are randomly sampled within a set of values at the beginning of the task. Once the path is represented by the spline, the waypoints are sampled at regular intervals of distance along the spline. The resulting samples are then used as reference positions for the end-effector, $\eta_{t}$, at each timestep. 
    
    The reward function for this task is calculated based on how closely the end-effector follows the trajectory:
    
    \begin{equation*}
    r(s_t) = -\alpha ||\eta_{t} - \mathcal{P}(t)||
    \end{equation*}
    
    where $\alpha$ is a scaling factor determining the penalty for deviating from the path and $\mathcal{P}(t)$ denotes the desired position on the path at time $t$.
    
    Supported environments for this task include the PSM \& ECM attachments for the da Vinci, and the STAR robot.

    \item \textbf{Multi-Tool Target Reaching:} In the Multi-Tool Target Reaching task, the robot is required to coordinate the movement and positioning of several end-effectors, (e.g. with $n=2$ there is $\eta_1$ and $\eta_2$) to align with multiple target spheres (e.g. $g_1$ and $g_2$). Each target sphere has its own distinct position, and the challenge lies in simultaneously positioning multiple tools and handling self-collisions. The reward function takes into account the distance between each end-effectors and their targets: 
    
    %\begin{equation*} 
    %r(s_t) = \rho[||\eta_{1} - g_{\eta_{1}}|| + ||\eta_{2} - g_{\eta_{2}}||]
    %\end{equation*} 

    \[ r(s_t) = \rho \sum_{i=1}^{n} ||\eta_i - g_{\eta_i}|| \]
    
    The environment setup is designed to simulate complex surgical scenarios where multiple tools need to be coordinated simultaneously. This task is an extension of the Target Reaching task and represents a higher level of complexity. Supported environments for this task include position two independent PSMs (Bimanual PSM), two indepdendent STARs (Bimanual STAR), and two PSMs together with one ECM (Trimanual PSM-ECM). 
\end{itemize}

\subsection*{Simulator Performance Comparison}

To properly assess the computational efficiency of various surgical robot learning platforms, we conducted performance profiling of each simulation environment. Time profiling begins after the first simulation step and ends after \textit{one million} simulation steps were reached. Profiling was done for each environment on their respective \textit{PSM Reach} task, which was a shared environment among all simulators. Simulation times were collected for each environment by running the simulator $30$ times in separate instances\footnote{Simulation speed times for LapGym and UnityFlexML were self-reported by the authors of the respective libraries.}. 

The performance of each simulator was first tested \textit{without} incorporating learning dynamics in order to evaluate the speed of the simulator by itself. In this condition, the physics simulator was given random action input ($a_i \sim U(-1, 1))$ and simulator feedback was ignored. Results are shown below in \textit{Table I}.

\begin{center}
\begin{table}[H]
\caption{Simulation speed of surgical simulators}
\centering
\begin{tabular}{ c c c } 
 \hline
 Simulator & Seconds per 1M Timesteps & Frames per second \\  
 \hline 
 UnityFlexML &  $17857 \pm 1123.1$ & $56 \pm 4.2$    \\ 
 UnityFlexML WD\footnote{Without deformable objects} &  $6410 \pm 124$ & $156 \pm 6.4$    \\ 
 LapGym &   $2166 \pm 52.1$ & $461 \pm 11$ \\
 dVRL & $2113 \pm 12.2$ & $473 \pm 3.1$ \\ 
 %AMBF-RL &   $? \pm ?$ & $? \pm ?$    \\ 
 SurRoL &  $242 \pm 3.2$ & $4124 \pm 54$ \\
 \hline
 \textbf{Surgical Gym} &  $\textbf{2.9} \pm \textbf{0.2}$ & $\textbf{344828} \pm \textbf{22247}$ \\ 
 \hline
\end{tabular}
\end{table} 
\end{center} 

As can be seen, Surgical Gym takes an average of $2.9$ seconds to collect $1M$ timesteps ($344828$ frames per second), whereas UnityFlexML takes as much as $17857$ seconds ($\sim 300$ minutes at $56$ frames per second). In terms of simulation time, Surgical Gym is $\sim7000 \times$ faster than the slowest simulator and $\sim80 \times$ faster than the fastest simulator.

Next, the performance was tested \textit{with} the incorporation of learning dynamics. In this condition, the simulator is given action input forward propagated through a neural network, reward is accumulated, and policy gradient updates are calculated. The neural network is a 3 hidden-layer network with hidden dimensions 256, 128, and 64 respectively using ELU nonlinear activations\cite{clevert2015fast}. Rather than testing the pure simulation speed, this test benchmarks how long it takes for the policy to be \textit{optimized} on one million simulation steps, validating the time efficiency of the \textit{learning dynamics}. Results are shown below in \textit{Table II}. %The number of neurons and hidden layers for the ANN were chosen to be consistent with the default provided the respective library (for Isaac Gym 3 hidden layers, with neuron counts 256, 128, 64 for respective layers).

\begin{center}
\begin{table}[H]
\caption{Simulation speed of simulator learning dynamics}
\centering
\begin{tabular}{ c c c } 
 \hline
 Simulator & Seconds per 1M Timesteps & Frames per second \\  
 \hline 
 UnityFlexML & $37037 \pm 1796$ & $27 \pm 2.9$ \\ 
 UnityFlexML WD\footnote{Without deformable objects} & $19231 \pm 1342$ & $52 \pm 3.9$ \\ 
 LapGym (8 core) & $833 \pm 13.5$ & $1200 \pm 19.1$ \\ 
 dVRL & $7024 \pm 26.5$ & $142 \pm 0.5$ \\
 %AMBF-RL &  $? \pm ?$ & ?    \\ 
 SurRoL & $612 \pm 35.2$ & $1634 \pm 94.3$ \\
 \hline
 \textbf{Surgical Gym} &  $\textbf{6.8} \pm \textbf{0.4}$ & $\textbf{147059} \pm \textbf{8170}$ \\ 
 \hline
\end{tabular}
\end{table}
\end{center}

Incorporating the learning dynamics, Surgical Gym takes an average of $6.8$ seconds to train on $1M$ timesteps ($147059$ frames per second), whereas UnityFlexML takes as much as $37037$ seconds ($\sim600$ minutes at $27$ frames per second). With respect to training times Surgical Gym is $\sim5500\times$ faster than the slowest simulator and $\sim90 \times$ faster than the fastest simulator.

For each of the libraries, default configurations were used for evaluation (e.g. LapGym using 8 cores). For Surgical Gym, we ran $20,000$ environments in parallel directly on a \textit{single} GPU ($8$GB NVIDIA Quadro RTX $4000$).

\section{Conclusion}

We see Surgical Gym expanding beyond the environments discussed in this paper. Future work could advance this library with more complex problems, such as in SurRoL \cite{xu2021surrol} and LapGym \cite{scheikl2023lapgym}. We also see the incorporation of tasks involving soft body models which more closely resemble surgical tasks, such as tissue manipulation \cite{li2020super}, which IsaacGym supports \cite{lim2022real2sim2real}. Transferring from simulation to hardware is more challenging with RL-trained torque and PD controllers than inverse dynamics (which was used in previous works). Thus, the further integration of V-REP kinematic models \cite{fontanelli2018v} redesigned with GPU-compatibility may make the transfer from simulation to hardware simpler. 

While Surgical Gym provides significant improvements in simulation speed, there is still much work to be done before this can be applied to autonomous surgery. We hope by making this work open-source, researchers can build on it to make sophisticated autonomous controllers for a wide variety of surgical applications (e.g. suturing, tissue manipulation, pattern cutting). Here, we introduce an open-source platform that accelerates surgical robot training with GPU-based simulations and RL. This platform features five training environments, with six surgical robots. We believe easier access to simulated training data will enable the next generation of autonomous surgery and hope Surgical Gym provides a step in that direction. 

\section*{Acknowledgement}

This material is based upon work supported by the National Science Foundation Graduate Research Fellowship for Comp/IS/Eng-Robotics under Grant No. DGE2139757 and NSF/FRR 2144348.

\vspace{8mm}

\bibliographystyle{ieeetr}

\bibliography{surgical_gym}

\end{document}